\documentclass{article}
\usepackage{spconf,amsmath,graphicx}
\usepackage{booktabs}

\usepackage{hyperref}
\title{Identifying every building's function in large-scale urban areas with multi-modality remote-sensing data}
\name{Zhuohong Li$^1$, Wei He$^1$, Jiepan Li$^1$, Hongyan Zhang$^{1, 2\dag}$ \thanks{\dag Corresponding author: zhanghongyan@cug.edu.cn}}
\address{\small $^1$State Key Laboratory of Information Engineering in Surveying, Mapping and Remote Sensing, Wuhan University, Wuhan 430079, China\\
\small $^2$School of Computer Science, China University of Geosciences, Wuhan 430074, China}

\usepackage{multirow}
\usepackage{enumitem}
\usepackage{subfigure}
\usepackage{hyperref}
\usepackage{caption} 
\begin{document}

\maketitle
\begin{abstract}
Buildings, as fundamental man-made structures in urban environments, serve as crucial indicators for understanding various city function zones. Rapid urbanization has raised an urgent need for efficiently surveying building footprints and functions. In this study, we proposed a semi-supervised framework to identify every building's function in large-scale urban areas with multi-modality remote-sensing data. In detail, optical images, building height, and nighttime-light data are collected to describe the morphological attributes of buildings. Then, the area of interest (AOI) and building masks from the volunteered geographic information (VGI) data are collected to form sparsely labeled samples.
Furthermore, the multi-modality data and weak labels are utilized to train a segmentation model with a semi-supervised strategy. Finally, results are evaluated by 20,000 validation points and statistical survey reports from the government. The evaluations reveal that the produced function maps achieve an OA of 82\% and Kappa of 71\% among 1,616,796 buildings in Shanghai, China. This study has the potential to support large-scale urban management and sustainable urban development. All collected data and produced maps are open access at \url{https://github.com/LiZhuoHong/BuildingMap}.  
\end{abstract}
\section{Introduction}

Mapping the footprints and functions of buildings, as basic applications of earth observation, provides an avenue to comprehend the urban layout \cite{bittner2018building,li2024learning}.
Over the past decades, the world has undergone a tremendous wave of urbanization. Large-scale, high-resolution (HR) building function maps can effectively facilitate urban planning and management \cite{bush2019building, naess2001urban}.

Due to the development of sensors and satellites, the available remote sensing data has undergone a transition from coarse to fine resolution \cite{wang2022cross,hu2023cross}. Abundant HR multi-modality data provides strong support for mapping the building function on a larger scale and with a finer gain.
However, current studies either focus on extracting the building footprint without function information or classifying the land parcels' functions, where the scale is not precise enough to identify each building. In detail, related studies can be summarized into the following two types:
\begin{figure}[]
{
    \begin{minipage}[b]{\hsize}
     \centering
    \includegraphics[width=0.95\linewidth]{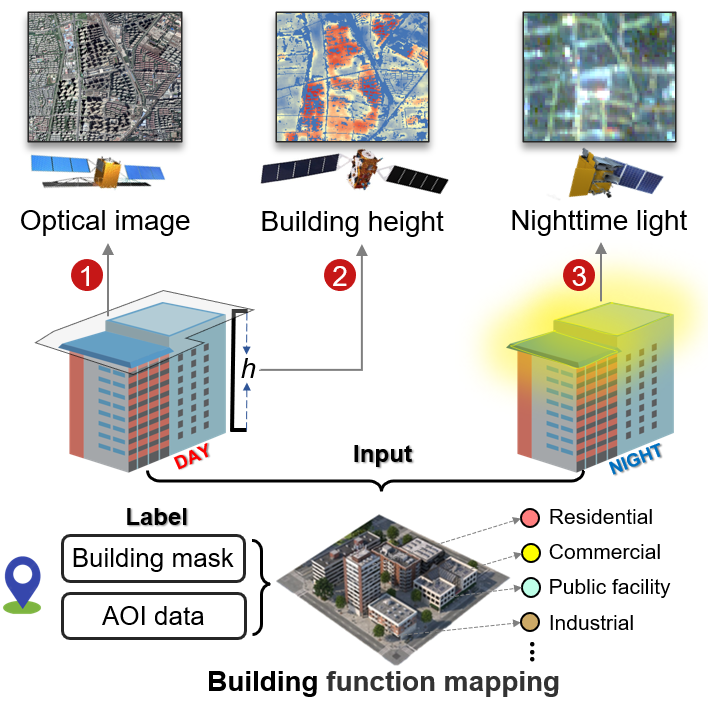}
    \end{minipage}
    } 
\vspace{-2em}
\caption{ \footnotesize\footnotesize \rmfamily Illustration of building function classification using multi-modality input (\textbf{Source}) and sparse VGI label (\textbf{Guide}) to generate HR building function maps (\textbf{Target}).}
\label{Motivation}
 \vspace{-1em}
\end{figure}
\\
\textbf{Building footprint mapping:} Extracting buildings from HR remote-sensing images has achieved great success \cite{zhang2020local,wang2022building}. The optical images describe the morphology (e.g., color, shape, and size) of the buildings and provide sufficient information for large-scale building footprint mapping \cite{he2023building}.
\begin{figure*}[t]
\centering
\includegraphics[width=0.95\linewidth]{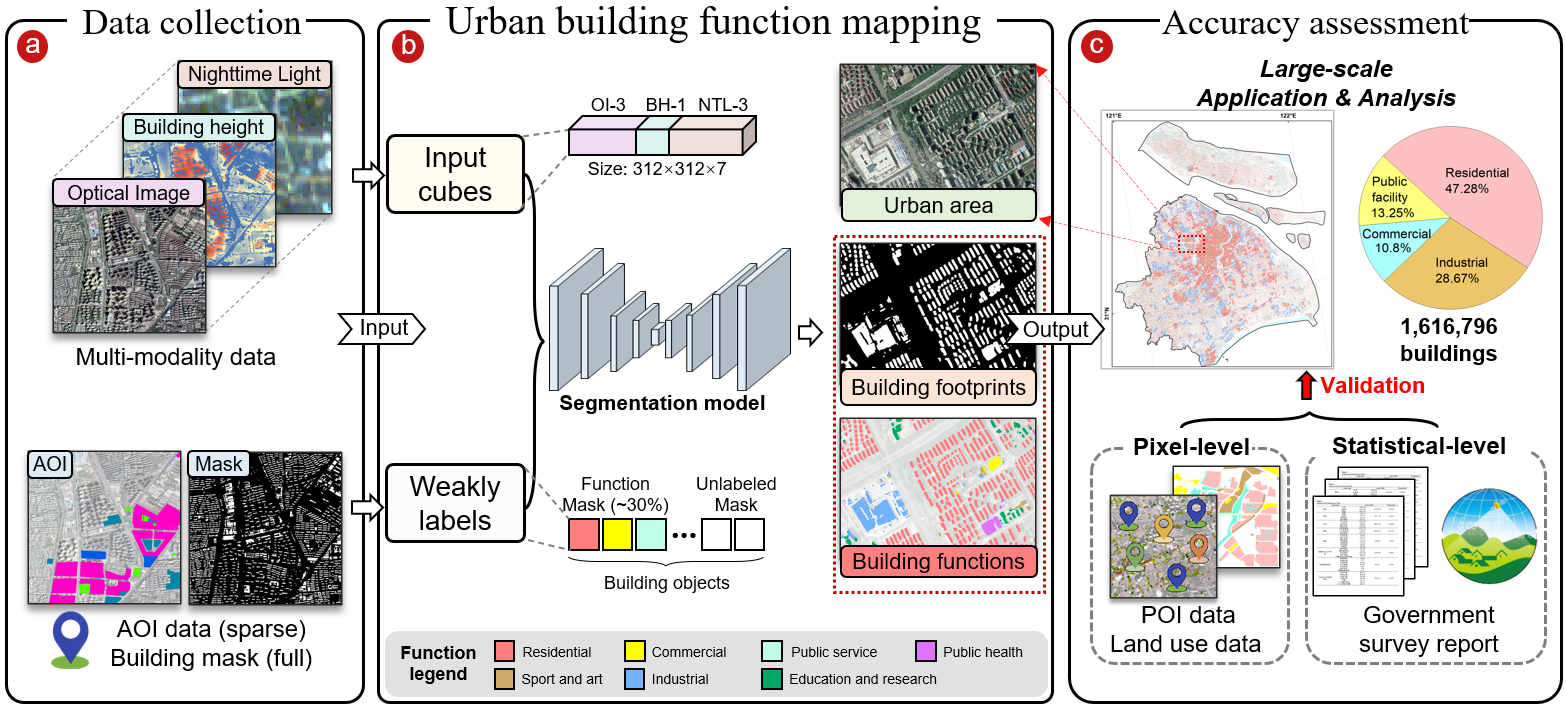}
\caption{ \footnotesize\rmfamily Overall workflow of the proposed framework. The framework includes three main parts: (a) Data collection; (b) Urban building function mapping; (c) Accuracy assessment. The optical images (OI), Building height (BH), and Nighttime-light data (NTL) contain three, one, and three bands, respectively.} 
\label{framework}
\vspace{-1em}
\end{figure*}
\\
\textbf{Land-use function mapping:} Studies of urban function zone investigation generally focus on parcel-scale analysis. The parcel divided by traffic routes is treated as the smallest unit for urban understanding and land-use mapping \cite{zhu2022knowledge, kuemmerle2013challenges}.

These studies emphasized extracting building footprints and classifying function zones separately. However, urban function zones are closely related to the function of buildings within the parcels. Current studies are commonly limited by single-mode data \cite{ding2021adversarial}, which is insufficient to describe and classify every building's function \cite{guo2020scene}. Moreover, creating accurate labels for a huge number of buildings is time-consuming and laborious \cite{li2022breaking,2021Outcome}, which raises another limitation for large-scale building function mapping.

In this study, we proposed a deep learning-based framework to automatically create large-scale building function maps using HR optical imagery, building height, nighttime-light data, and volunteered geographic information (VGI) as input data. In detail, the multi-modality data shown in Fig. \ref{Motivation} are integrated as the input cubes. The area of interest (AOI) data and building masks are integrated to generate labels with sparse supervised information. The input cubes and weak labels are used to train a segmentation model with a semi-supervised strategy. Then, the building footprints and function maps are produced by the well-trained network simultaneously. To comprehensively evaluate the produced building maps, we collected over 20,000 points for pixel-level validation. Additionally, statistical-level validation sets were collected from the government survey report.

\begin{figure*}[t]
\centering
\includegraphics[width=\linewidth]{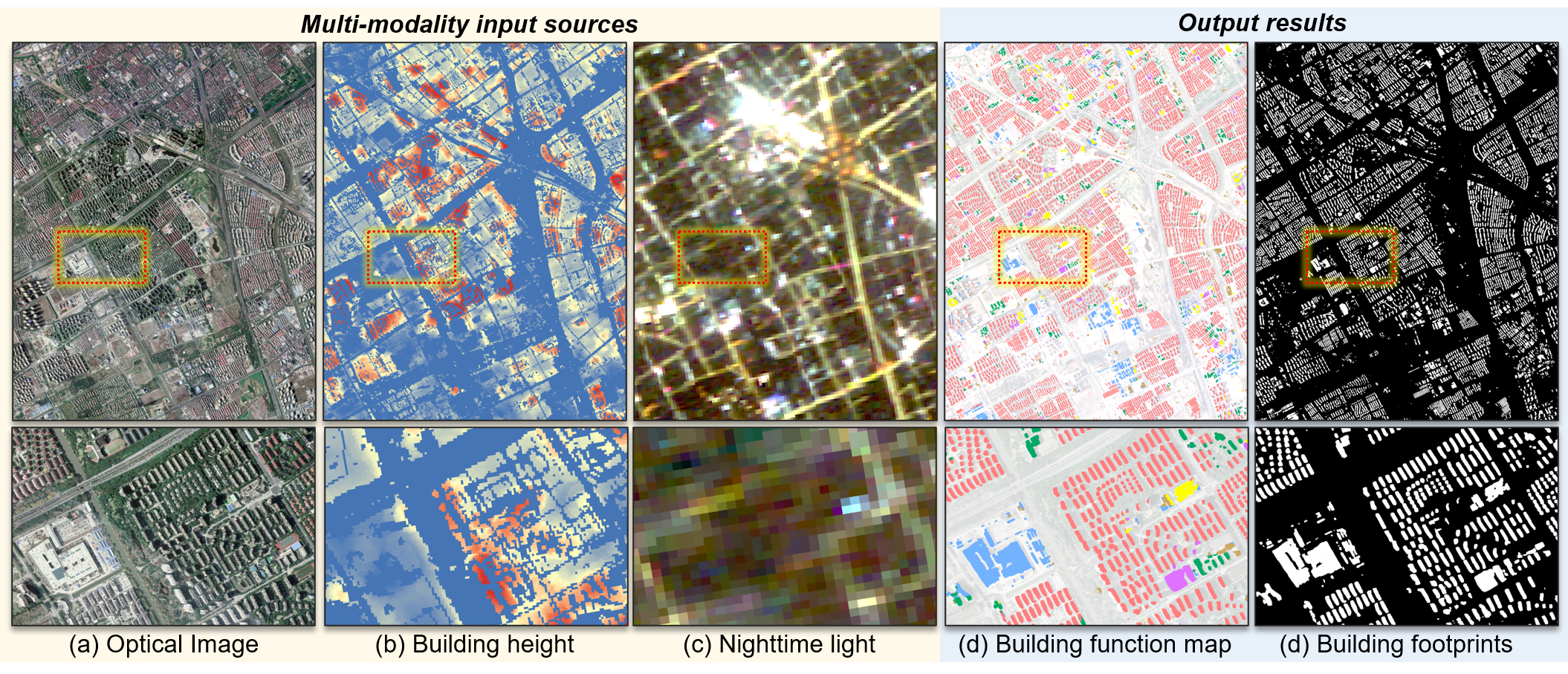}
\vspace{-2em}
\caption{ \footnotesize\rmfamily Demonstration of the multi-modality data, the produced building function maps, and corresponding building footprints.} 
\label{visual_results}
\vspace{-1em}
\end{figure*}
\vspace{-1em}
\section{methodology}
\vspace{-1em}
In this section, we sequentially present the processes of data collection and building function mapping. The accuracy evaluation will be discussed in the experimental section.
\vspace{-1.5em}
\subsection{Data collection}
\vspace{-0.5em}
To collect reliable training data, as shown in Fig.\ref{framework} (a), three data modalities are collected to form input cubes, and two types of VGI data are collected to generate training labels. 
\textbf{Study area} covers the whole of Shanghai City, one of the largest cities and international financial centers in China, covering 6340.5 $km^2$.\\
\textbf{Input cubes} are organized into 272 non-overlapping tiles, each with a size of 6000 × 6000 pixels. The specific data includes:
\begin{enumerate}
\vspace{-0.5em}
\small
\item \emph{The HR optical images (1 m/pixel)} are collected from the Google Earth imagery. The images contained three bands of red, green, and blue \cite{li2023sinolc}.
\vspace{-0.5em}
\item \emph{The building height (10 m/pixel)} is from the Chinese building height estimate dataset (CNBH-10 m) \cite{wu2023first}.
\vspace{-0.5em}
\item \emph{The nighttime-light data (10 m/pixel)} are from the SDGSAT-1, the world’s first scientific satellite for sustainable development goals \cite{guo2023sdgsat}.
\end{enumerate}
\vspace{-0.5em}
\textbf{Weak labels} are generated by conducting logical operations on the following AOI data and building masks:
\begin{enumerate}
\small
\vspace{-0.5em}
\item \emph{The AOI data (vector)} are collected from the OSM data. The AOI data contains more than 100 building function types (e.g., cinema, hospital, department, and market) \cite{li2023sinolc}.
\vspace{-0.5em}
\item \emph{The building masks (vector)} are from the building rooftop dataset containing 90 Chinese cities' building masks \cite{zhang2022vectorized}.
\end{enumerate}
\vspace{-1.5em}
\subsection{Urban building function mapping}
\vspace{-0.5em}
To generate the training pairs, as shown in Fig.\ref{framework} (b), the multi-modality data are concatenated to form the input cubes with seven bands. The segmentation model randomly crops 200 cube patches ($312 \times 312 \times 7$) from one tile ($6000 \times 6000 \times 7$). Based on practical demand, we unify more than 100 classes of AOI data into seven types of building functions.

To comprehensively utilize the multi-modality data, we adopt the high-resolution network (HRNet) as the segmentation model to extract the various, dense building attributes and classify the building functions. HRNet is constructed by parallel multi-scale convolutional branches to prevent the HR features from being down-sampled. Then, HRNet fuses the multi-scale features of these branches to enhance the semantic representation \cite{wang2020deep}. Based on the model, the fine edges and details of HR images can be well extracted and preserved. Meanwhile, the building height and nighttime-light data can synthetically enhance the building function identification.

Moreover, the AOI data only covers about 30\% area of the urban \cite{senaratne2017review}, while the building masks entirely cover the area. Therefore, the footprint labels include every building object within the study area, but only a small number of them have function-labeled information.
Based on weak labels, we adopt a semi-supervised strategy for the training process of HRNet. In detail, we modify the cross-entropy (CE) loss by considering the background and all function types as the supervised classes. Then, building objects without function-labeled information are regarded as unsupervised classes, which are ignored during loss calculation. Formally, for a training patch of the size of $W \times H$, we used ${\bf{Y'}}$,${\bf{\hat Y}}$, and $\bf{G}$ to represent the labels, results, and the unlabeled building mask, respectively. The modified CE loss can be written as:
\begin{equation} 
\small
{\mathcal{L}_{MCE}}({\bf{Y'}},{\bf{\hat Y}},{\bf{G}}) = \frac{{ - \sum\limits_{i=0}^W {\sum\limits_{j=0}^H {\left[ {{{ g}_{ij}}\sum\limits_{l = 1}^L {{{y'}_{ij}}^{(l)}\log ({{\hat y}_{ij}}^{(l)})} } \right]} } }}{{{\rm{Sum}}({\bf{G}}(i,j) = 1)}}.
\label{CAS_LOSS}
\end{equation}

In Eq. \ref{CAS_LOSS}, ${y'_{ij}}^{(l)}$ and ${\hat y_{ij}}^{(l)}$ denote the pixel $(i,j)$ of the label ${\mathbf{Y'}}$ and prediction ${\mathbf{\hat Y}}$ with the class $l$. If the pixel is a building object without function-labeled information, the ${g}_{ij}$ is set to 0, otherwise the ${g}_{ij}$ is set to 1.
\begin{figure}[!h]
\subfigure[Qualitative results of Shanghai. Three mapping samples are demonstrated.]{
    \begin{minipage}[b]{\hsize}
     \centering
    \includegraphics[width=\linewidth]{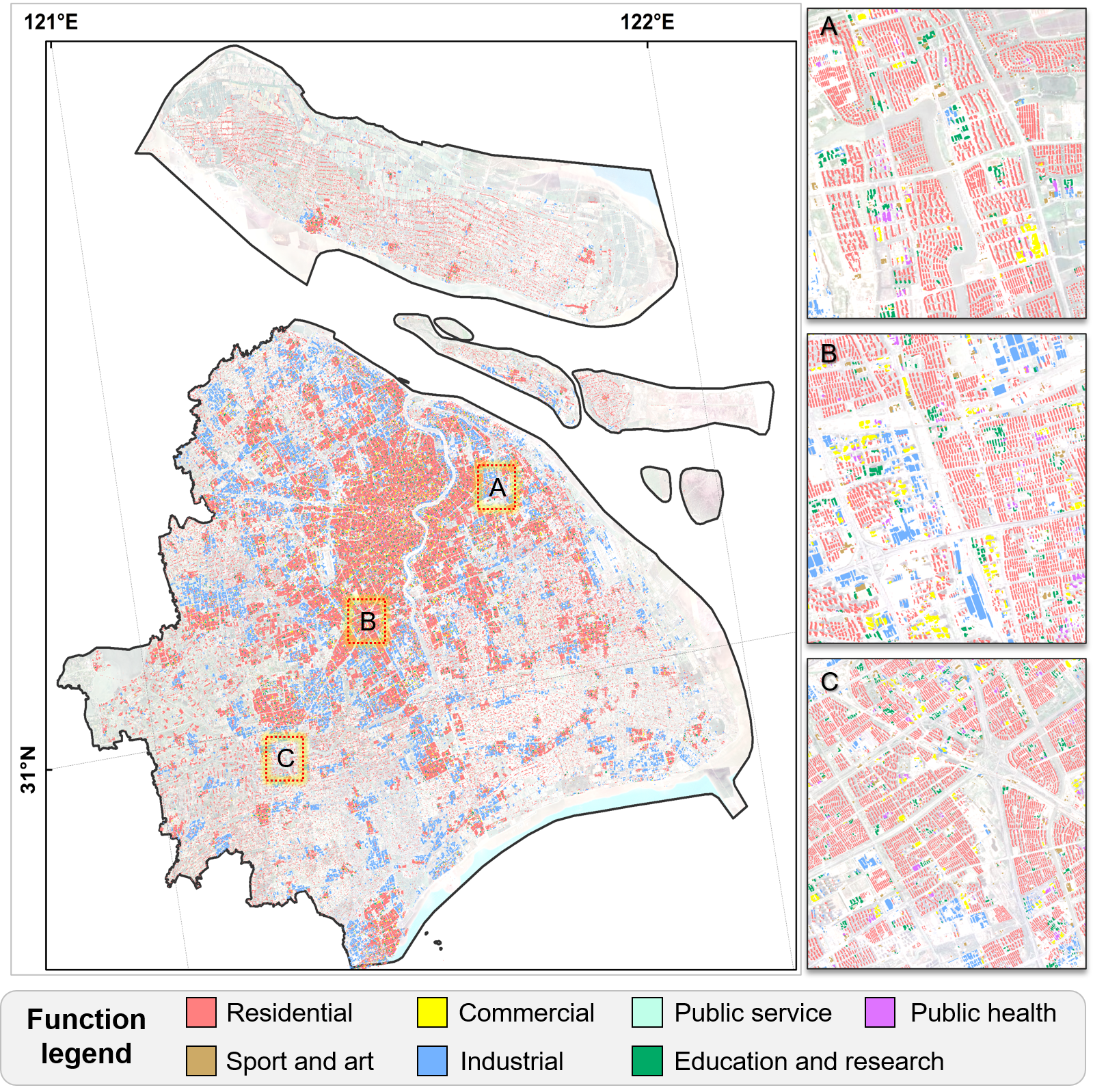}
    \end{minipage}
    } \\
    \subfigure[Average building heights and nighttime-light intensities among different building function types.]{\hspace{-2mm}
    \begin{minipage}[b]{\hsize}
     \centering
    \includegraphics[width=\linewidth]{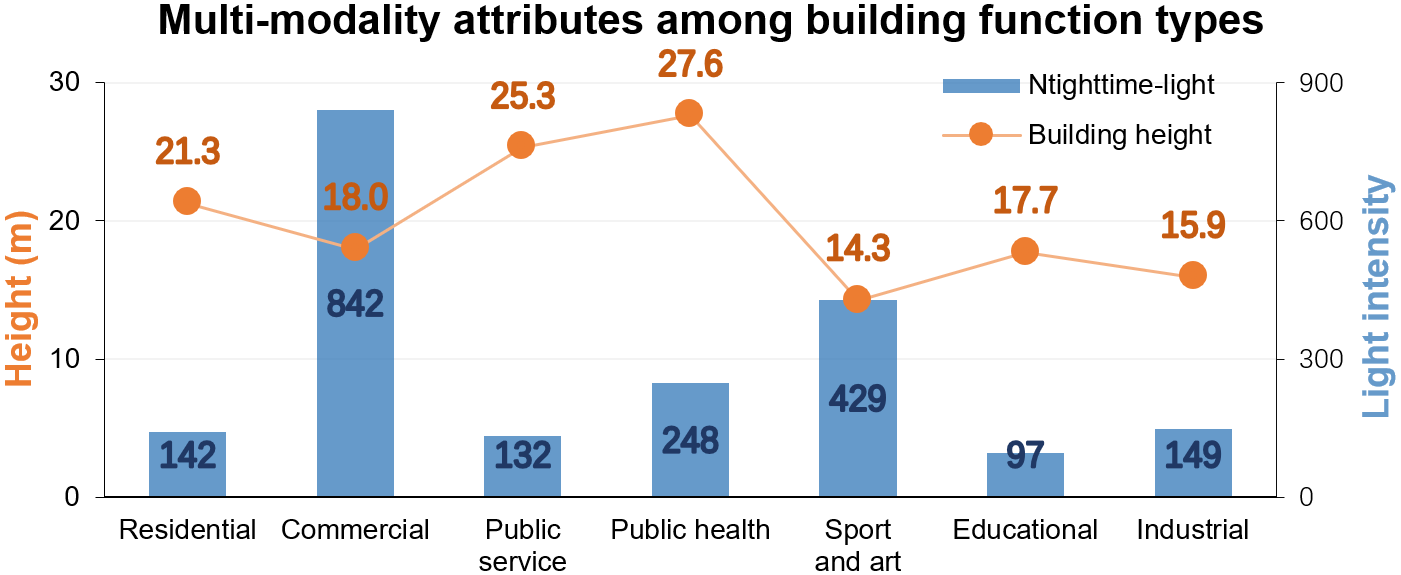}
    \end{minipage}
    }
        \vspace{-1em}
\caption{ \footnotesize\footnotesize \rmfamily Demonstration of the overall results and evaluation in the study area.}
\label{overall visual}
    \vspace{-2em}
\end{figure}

    \vspace{-1em}
\section{Experiments}
\vspace{-0.5em}
To validate the results, as shown in Fig. \ref{framework} (c), we collected $>20,000$ validation points across the study area. Moreover, we derived a statistical validation set by collecting the National Urban Construction Land Report from the Ministry of Housing and Urban-Rural Development of the government\footnote{https://www.mohurd.gov.cn/}. 

\vspace{-1em}
\subsection{Qualitative analysis}
\begin{figure}[]
{
    \begin{minipage}[b]{\hsize}
     \centering

    \includegraphics[width=\linewidth]{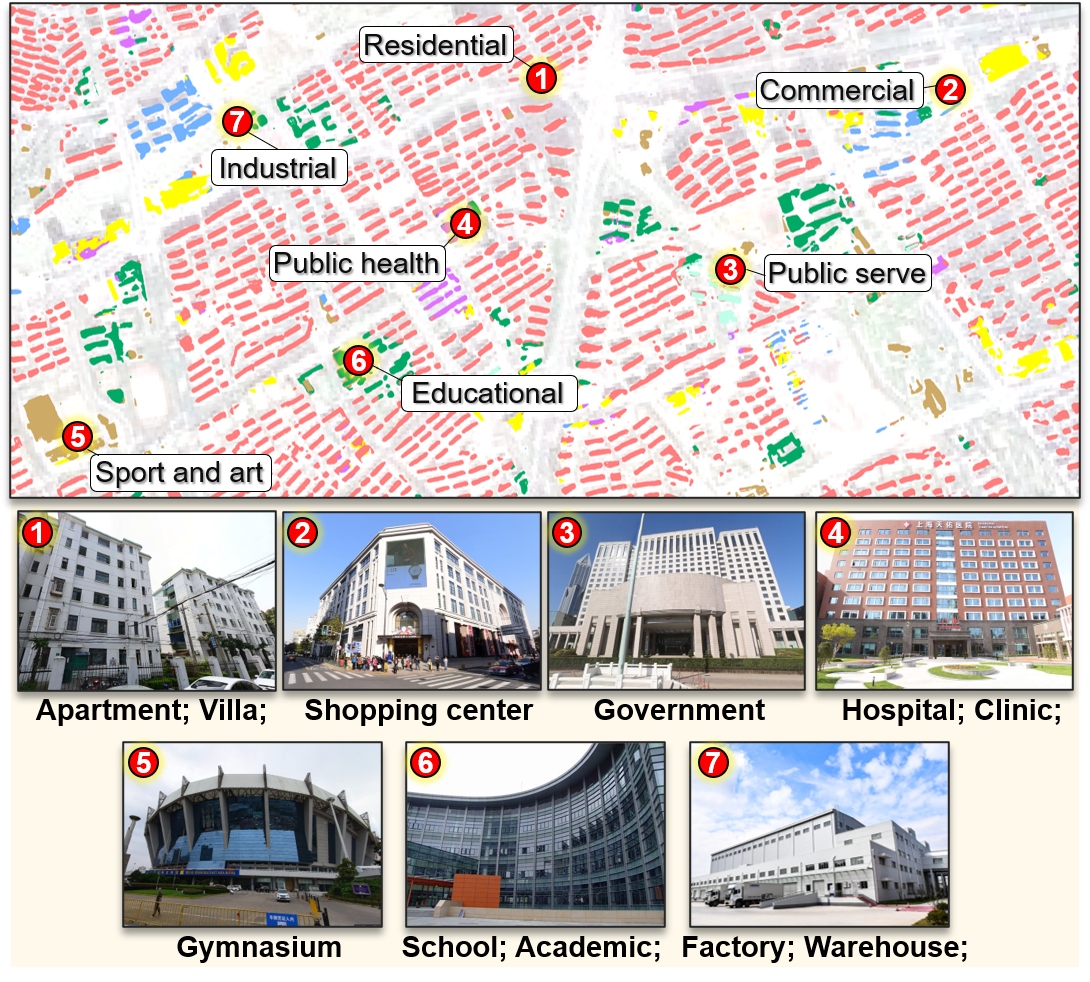}
    \end{minipage}
    } 
    \vspace{-2em}
\caption{ \footnotesize\footnotesize \rmfamily Pattern analysis and street views of typical buildings in Shanghai.}
\label{sample}
    \vspace{-2em}
\end{figure}

\vspace{-0.5em}
Fig. \ref{visual_results} shows the multi-modality data and the identified building functions and footprints. The results reveal that buildings with different functions have distinctive features among multi-modality data. In detail, as shown in Fig. \ref{overall visual} (b), residential buildings are generally higher with weaker nighttime light, but commercial buildings are lower with the strongest nighttime light.
Additionally, Fig. \ref{overall visual} (a) shows that buildings within the overall urban of Shanghai have been well identified. Fig.\ref{sample} demonstrates the produced building function maps and online street views \cite{yu2022spatio} of typical building types, which indicated a clear city pattern and function zone division.
\vspace{-2.5em}
\subsection{Quantitative analysis}
\begin{figure}[]
{
    \begin{minipage}[b]{\hsize}
     \centering
        \includegraphics[width=\linewidth]{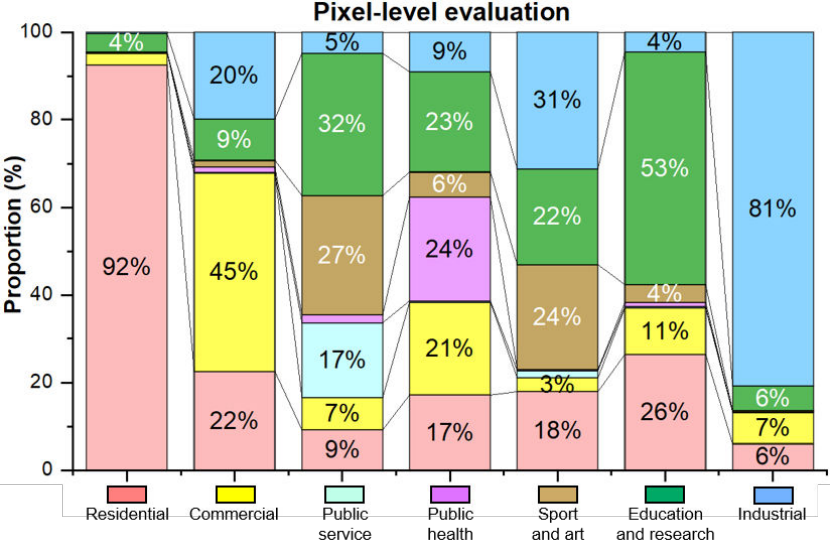}
    \end{minipage}
    } 
    \vspace{-2em}
\caption{ \footnotesize\footnotesize \rmfamily Confusion proportions for each building function type.}
\label{matr}
    \vspace{-0em}
\end{figure}

\begin{figure}[]
{
    \begin{minipage}[b]{\hsize}
     \centering

    \includegraphics[width=\linewidth]{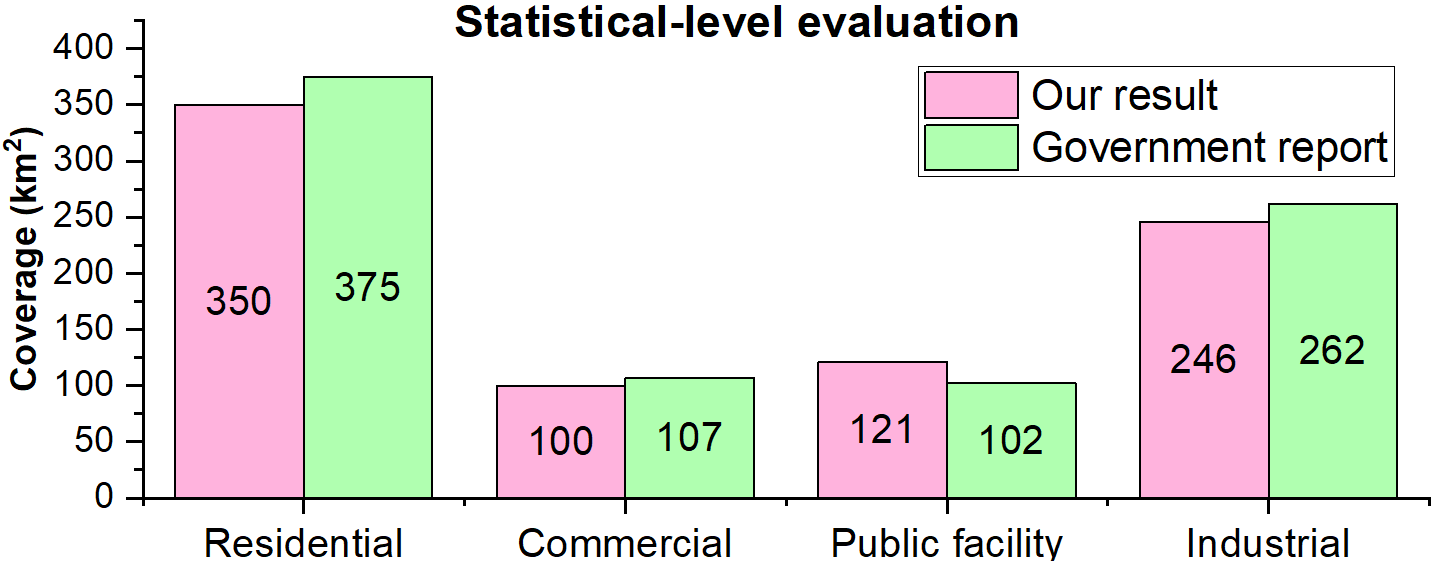}
    \end{minipage}
    } 
    \vspace{-2em}
\caption{ \footnotesize\footnotesize \rmfamily Statistical-level comparison between results and official reports.}
\label{sta}

\end{figure}

\vspace{-1em}
Based on the pixel-level validation, Fig. \ref{matr} shows the confusion proportions for each building type. The evaluation indicates that residential, industrial, educational, and commercial buildings are classified more accurately. Public services and public health have the lowest proportion and insignificant attributes, resulting in lower accuracy.
Based on the statistical-level validation, Fig. \ref{sta} shows the comparison between the results and government reports. The results are unified into four basic classes, according to the office reports. The comparison indicates the produced function maps are generally consistent with the authoritative report.
Table \ref{overall table} shows the overall evaluation results. The residential, industrial, commercial, and educational buildings occupied the largest area of Shanghai. For the building function maps, the results obtain an overall accuracy (OA) of 81.81\%, a Kappa of 70.60\%, and a frequency-weighted intersection over Union (FWIoU) of 72.80\%. For the building footprints, an F1 score of 85.73\% and an IoU of 75.03\% are obtained, which indicates 1,616,796 buildings in Shanghai with a total coverage of 820.31 $km^2$. In general, the study provides a complete survey and visual maps for building distribution and composition in the urban.

\begin{table}[!h]
\small
\caption{\footnotesize Overall quantitative evaluation and building function analysis.}
\vspace{-1em}
\begin{tabular}{ll|ll}
\hline

Function type  & Proportion & \multicolumn{2}{l}{Overall evaluation} \\ \hline
Residential    & 50.06\%    & OA                   & 0.8181         \\
Commercial     & 4.82\%     & Kappa                & 0.7060         \\
Public service & 0.43\%     & FWIoU                & 0.7280         \\\cline{3-4} 
Public health  & 1.76\%     & Footprint IoU        & 0.7503         \\ 
Sport and art  & 2.85\%     & Footprint F1         & 0.8573         \\
Educational      & 4.84\%     & Building count       & 1,616,796         \\
Industrial     & 35.24\%    & Building area        & 820.31 $km^2$      \\ \hline
\end{tabular}
\label{overall table}
    \vspace{-1em}
\end{table}
\section{Conclusion}
\vspace{-0.5em}
In this study, we have developed a deep learning-based building mapping framework that predicts the function and footprint of every building in large-scale urban areas.
The qualitative analysis reveals that buildings with different functions can be effectively distinguished using multi-modality remote sensing data, and the generated maps clearly depict the city layout of Shanghai.
The quantitative validations indicate that the produced function maps achieve an OA of 82\% and a Kappa of 71\%. The results also statistically conform to the official survey reports from the government. In addition, a total of 1,616,796 building footprints in Shanghai are detected with an IoU of 75\%.
With the fine-grained building information, the proposed framework shows great potential to support urban management and ultimately to help cities move towards sustainable development goals.
In the future, we will continuously improve the classification of easily confused building types with more related data and process the national-scale building function mapping for all major cities in China.


\small
\bibliographystyle{IEEEbib}
\bibliography{strings,refs}

\end{document}